\documentclass[11pt]{article}
\usepackage{lmodern}  
\usepackage{ACL2023}

\usepackage{times}
\usepackage{latexsym}
\usepackage[colorinlistoftodos,prependcaption,textsize=tiny]{todonotes}

\usepackage{xspace,mfirstuc,tabulary}
\usepackage{booktabs}

\usepackage{mdframed}
\usepackage{times,latexsym}
\usepackage{url}
\usepackage[notransparent]{svg}
\usepackage{amsmath}
\usepackage{graphicx}
\usepackage[T1]{fontenc}
\usepackage{microtype}
\usepackage{amsmath}
\usepackage{tabularray}
\usepackage{algorithm2e}
\usepackage{comment}

\usepackage[T1]{fontenc}

\usepackage[utf8]{inputenc}

\usepackage{microtype}

\usepackage{inconsolata}

\author{
Guillem Ramírez$^1$   \and Matthias Lindemann$^1$  \and Alexandra Birch$^1$  \and Ivan Titov$^{1,2}$  \\
$^1$ ILCC, University of Edinburgh,
$^2$ ILLC, University of Amsterdam \\
\texttt{gramirez@ed.ac.uk}
}

\title{Cache \& Distil: Optimising API Calls to Large Language Models}

\date{}

\begin{document}
\maketitle
\begin{abstract}
Large-scale deployment of generative AI tools often depends on costly API calls to a Large Language Model (LLM) to fulfil user queries. To curtail the frequency of these calls, one can employ a smaller language model -- a \emph{student} -- which is continuously trained on the responses of the LLM. This student gradually gains proficiency in independently handling an increasing number of user requests, a process we term \emph{neural caching}. The crucial element in neural caching is a policy that decides which requests should be processed by the student alone and which should be redirected to the LLM, subsequently aiding the student’s learning. In this study, we focus on classification tasks, and we consider a range of classic active learning-based selection criteria as the policy. Our experiments suggest that Margin Sampling and Query by Committee bring consistent benefits across tasks and budgets.

\end{abstract}

\section{Introduction}
Large Language Models (LLMs) offer unique capabilities in understanding and generating human-like text. They have become indispensable in a wide range of applications, including assistive tools and entertainment bots. However, large models are often very challenging for all but a few companies and institutions to run on their infrastructure~\cite{DBLP:journals/cacm/SchwartzDSE20}. Meanwhile, smaller models typically under-perform in these applications, at least without additional fine-tuning on task-specific labelled data. Consequently, many applications access LLMs via commercial APIs despite the costs involved and the exposure of their entire request stream to the API providers.

\begin{figure}
  \centering
\includegraphics[trim={6cm 20.3cm 3cm 1cm},clip, width=\linewidth]{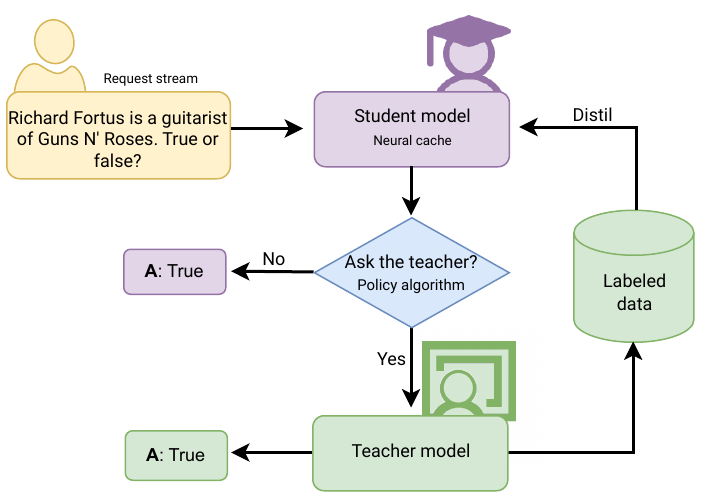}
    \caption{Neural caching (one iteration): A student generates a response to a user request. The policy algorithm determines whether to rely on the student's response or to call an LLM. LLM responses are stored and used to re-train the student as more data becomes available.}
    \label{diagram}
\end{figure}

To minimise the costs and data exposure associated with calling the API, it is natural to consider a scenario where a smaller language model, which we refer to as \emph{student}, is being trained on the LLM's predictions and, as the student gets more accurate, it handles an increasing number of requests. The knowledge of the LLM gets continuously distilled into the smaller model. We refer to this scenario as \textit{neural caching} (see Figure~\ref{diagram}), as the student can be thought of as a smart cache, remembering what the LLM predicted in the past. Neural caching can be regarded as a more powerful version of look-up tables, previously used to cache LLM predictions \citep{GPTCache,zhu_optimal_2023}. The goal of this paper is to formalise the neural caching problem and investigate simple ways to approach it. 

The key element in the neural caching scenario is the policy determining which requests the student processes independently. A good policy should weigh the expected immediate user benefit (i.e., if the LLM is substantially more likely to make a correct prediction than the student) and the anticipated benefit for the student (i.e., whether the LLM's prediction will aid in training the student). 
The latter underscores its relationship with active learning~\citep[AL,][]{settles_active_2009, ouyang_training_2022}, although AL is typically associated with soliciting human annotations. In particular, there is a similarity to online AL~\citep{cacciarelli_survey_2023}, where new unlabelled data points arrive in a stream and are discarded immediately or sent to an annotator. However, online AL tends to focus on maximising the accuracy of the final model (i.e. student in our terminology).
In contrast, what matters in neural caching is the accuracy of the joint system (student, teacher, along with the policy) over its lifetime since this \emph{online accuracy} reflects the average level of service offered to a user.

Despite the aforementioned differences with AL, evaluating the existing AL algorithms -- specifically the example selection criteria -- remains valuable given the maturity of the AL field and the ease of implementation of some of the AL methods. This study aims to achieve this, as well as to investigate the potential shortcomings of these methods. For instance, will the AL methods end up selecting examples that are too challenging even for the LLM? Would introducing these noisy examples be detrimental to the student? Answering these questions can inform future research on this practically significant scenario.

In this work, our focus is specifically on classification tasks, as opposed to free text generation. Many practical problems, such as routing user requests to relevant departments or answering questions about factual knowledge, can be framed as classification tasks. By confining our focus to classification, we can apply methods developed in AL without modification. This also allows us to circumvent additional challenges tied to the automatic evaluation of text generation~\cite{DBLP:journals/corr/abs-2006-14799}.  

Our findings reveal the benefits of using AL-based policies such as Margin Sampling~\cite{scheffer_active_2001} and Query by Committee~\cite{seung_query_1992}. Across datasets and budgets, these methods consistently outperform baselines, such as routing examples randomly or training the student at the very start.
Our analysis also reveals that the student appears robust to the noise introduced by an LLM, suggesting that the noise introduced by LLMs (especially on harder examples) does not influence them as much as one may expect. We also analyse a simplified practical scenario where the student is not retrained and observe even greater improvements in online accuracy from using AL-based policies. We release our code to encourage further work on this problem.\footnote{\href{https://github.com/guillemram97/neural-caching}{https://github.com/guillemram97/neural-caching}}
\newpage
The key contributions of this work are as follows:
\begin{itemize}
\item We formulate the \textit{neural caching} problem as a powerful extension of using static caches. In neural caching, LLM calls are optimised, while the student model is periodically retrained on the labels. This is, to our knowledge, the first work that leverages online knowledge distillation, which we believe could play a key role in saving unnecessary calls to expensive models.

\item We release a benchmark with LLM annotations for classification tasks to facilitate future research.

\item We evaluate and analyse different instance selection criteria for the neural caching setup.

\item Our findings reveal that AL-based selection criteria consistently improve performance over baseline methods across various budgets and datasets.

\end{itemize}

\section{Related Work}

\paragraph{Active Learning.}
AL seeks to reduce the amount of manual data annotation needed. To accomplish this, it selects the most informative examples from unannotated data. These datapoints are then presented to an annotator and the labels are subsequently used to train a model. The most common scenario for AL is pool-based, where a large unlabelled dataset is available at the beginning and then a subset of examples is selected for labelling. There has been extensive work on applying pool-based techniques to NLP tasks, especially for classification problems~\citep{settles_active_2009, ouyang_training_2022, zhang_survey_2023}.  

\paragraph{Online Active Learning.} In single-pass online AL~\citep{cacciarelli_survey_2023}, access to a large unlabelled dataset is not available. Instead, we are given one unlabelled instance at a time and need to decide at that time whether to request annotation.  
Online AL was initially motivated by scenarios in which an instance would not be available for annotation at a later time, such as in defect detection or medical applications, where an item might get shipped or the patient becomes unavailable~\citep{riquelme2017}. %
Online AL tends to focus on the final accuracy of the model, rather than the online accuracy of the combined system of student and teacher, the measure more suitable for our scenario. 

\paragraph{Knowledge Distillation of LLMs.} Knowledge distillation (KD), i.e., training a smaller model to mimic a larger one, has garnered substantial attention~\cite{DBLP:conf/kdd/BucilaCN06,DBLP:journals/corr/HintonVD15}. The class of methods most closely related to ours is active KD, which effectively applies AL to KD~\cite{DBLP:conf/iclr/LiangHSZCCC21,DBLP:journals/pr/XuLL23,baykal_robust_2023}. Similar to AL, the emphasis is placed on the pool-based setting, as opposed to the online setting, with a particular focus on optimising the final accuracy of the student model, rather than online accuracy as needed for our use case.

\paragraph{Optimisation of Commercial LLM API Calls.} 
Due to the high cost of commercial LLM APIs, several works have explored methods to reduce or otherwise optimise the cost of API calls.
GPTCache \citep{GPTCache} relies on a vector store of past query embeddings and retrieves their associated labels. It shares similarities with the Coreset version of our approach -- which emerged as the weakest method in our experiments. FrugalGPT~\citep{chen_frugalgpt_2023} implements a cascade of commercial LLMs, where bigger models are only called if the response from a cheaper model is deemed as too unreliable by a scorer that was trained with in-domain data. In contrast, our method does not assume we have readily available gold data to train a scorer.  
\citet{zhu_optimal_2023} present a method to allocate queries 
among multiple models, together with traditional caching, in a scenario with highly repetitive queries. \citet{sakota_fly-swat_2023, shnitzer2023large} optimise routing calls through models by predicting their respective performance. Our work deviates from all these as we propose to use continuous KD in a student model. 
\section{The Neural Caching Problem}
\label{caching}
The objective of neural caching is to optimise the usage of an LLM in a scenario where labels need to be generated for a stream of inputs. As we get more predictions from the LLM, a student model is trained on them. Our goal is to achieve the highest level of service possible within a set budget of LLM calls; hence, calling the LLM serves both to attain high accuracy for the incoming input as well as to train a student model.

To put it formally, our goal is to establish a mapping between elements in the input space $\mathcal{X}$ and the corresponding labels in the space $\mathcal{Y}$. We start with a student model $\mathcal{S}_0$, and we can access a teacher model $\mathcal{T}$ on demand. Our task is to predict labels for a sequence of $n$ examples ($x_1, \dots, x_n$) $\overset{\mathrm{iid}}{\sim}$ $\mathcal{X}$.  

We retrain the student model on the labels obtained from the LLM every $f$ processed requests. This simulates the situation where the number of requests is uniform in time, and there is a set time to retrain the model, e.g.\ at night. For simplicity and to follow the convention in AL to retrain the model from scratch~\citep{DBLP:journals/csur/RenXCHLGCW22}, every time we retrain the student model, we reset it to the original pre-trained model and then use parameter-efficient fine-tuning. Although continual learning methods could be employed~\cite{DBLP:conf/coling/BiesialskaBC20,DBLP:conf/aaai/0002C21}, we believe this is largely orthogonal to our primary focus on policies and resetting enhances the reproducibility of our analysis. Importantly, we do not assume access to ground truth (or human annotation) at any point in learning or in the calibration of the policy to simulate a fully automatic scenario.

For every new input $x_i$, we use the student model $\mathcal{S}_{i/f}$ to obtain the predicted label $\hat{y}^{S}_i$. Then, we have the option to request the label $\hat{y}^{\mathcal{T}}_i$ from the teacher model (LLM), which incurs a cost of $c(x_i)$. Finally, we return the label $\hat{y}_i$ for $x_i$: the teacher's label if requested or the student's otherwise. 

The processing of the $n$ examples is subject to a budget constraint, where the total cost must not exceed a fixed budget $b$. We assess the effectiveness of our querying strategy based on the accuracy of our predicted label $\hat{y}_i$ compared to the actual label $y_i$ (\textit{online accuracy}) on the online examples. Additionally, we measure the accuracy of the final student model $\mathcal{S}_{n/f}$ on a test dataset (\textit{final accuracy}). Algorithm~\ref{pseudocode} describes the process.

\RestyleAlgo{ruled}
\begin{algorithm}
\caption{Pseudo-code for the neural caching algorithm with budget $b$, retraining frequency $f$, cost per query $c$, data from the LLM $\mathcal{D}_{\text{LLM}}$ and an initial student $\mathcal{S}_{0}$}\label{pseudocode}
$\mathcal{D}_{\text{online}} = \emptyset$ \\
\For{$x_i$ \textnormal{\textbf{in} $X_{\text{online}}$} }
{
  \If{$i \bmod f$ == 0}{
    $\mathcal{S}_{i/f} = \text{Train}(\mathcal{D}_{\text{LLM}})$ \\ 
    }
  $\hat{y}_i = \mathcal{S}_{i/f} (x_i)$\\
  \If{\textnormal{Call\_LLM($b$, $x_i$, $\hat{y}_i$) \textbf{and} $b \geq c(x_i)$}}
  {
    $\hat{y}_i = \text{LLM}(x_i)$ \\
    $b = b - c(x_i)$ \\ 
 
$\mathcal{D}_{\text{LLM}} = \mathcal{D}_{\text{LLM}} \cup \{ \langle x_i, \hat{y}_i \rangle \} $\\
  } 
 
  $\mathcal{D}_{\text{online}} = \mathcal{D}_{\text{online}} \cup \{ \langle x_i, \hat{y}_i \rangle \} $\\
} 
$\mathcal{D}_{\text{test}} = \{\langle x_j, \mathcal{S}_{i/f} (x_j) \rangle \mid x_j \in X_{\text{test}}\}$ \\
$\text{Acc}_{\text{online}}= \text{Evaluate} (\mathcal{D}_{\text{online}})$ \\
$\text{Acc}_{\text{final}}= \text{Evaluate}(\mathcal{D}_{\text{test}})$
\end{algorithm}

\subsection{Instance Selection Criteria}
\label{strat}
We use classical instance selection criteria from AL for the neural caching problem. We use the term \textit{selecting an instance} to denote using the LLM to annotate that example.
\paragraph{Front-loading (FR)} This simple approach involves using the entire budget initially by selecting all instances for LLM annotation. Once the budget is used up, subsequent requests are handled by the student model alone. As in our experiments, the examples are i.i.d.; this strategy has the same expected \textit{final accuracy} as random selection.
\paragraph{Margin Sampling (MS)} MS~\citep{scheffer_active_2001, luo_active_2004} selects examples with high margin between the top two predictions made by the student model
\begin{equation}
\begin{aligned}
  \text{Margin}(x_i) =  &
  \quad \log P(y_i = k_1^\ast \mid x_i) \\ &  - \log P(y_i = k_2^\ast \mid x_i ) 
\end{aligned}
\label{BT}
\end{equation}
\noindent where $k_1^\ast$ and $k_2^\ast$ are the first and second most likely labels, respectively, according to the distribution $P(y_i \mid x_i)$ computed by the student model. This is a popular selection criterion for AL~\citep{roth_margin-based_2006, balcan_margin_2007}. 
\citet{schroder_revisiting_2022} evaluated different uncertainty-based strategies with Transformer models~\citep{devlin_bert_2019} and found MS to be the best-performing one in an offline, pool-based setting. To adapt MS -- as well as the other criteria -- to an online setting as a selection policy, we define a threshold, and only examples with a margin above this threshold are selected until the budget is exhausted. We refer to Appendix~\ref{thresholds} for more details.

\paragraph{Prediction Entropy (PE)} In PE~\citep{schohn_less_2000,roy_toward_2001}, we select instances with high entropy of the output distribution:
\begin{equation}
\begin{aligned}
  & \text{Entropy}(x_i) = \\
  & -\sum_j P(y_i = k_j^\ast \mid x_i) \log P(y_i = k_j^\ast \mid x_i) 
\end{aligned}
\end{equation}

\paragraph{Query by Committee (QBC)} In QBC~\citep{seung_query_1992, burbidge_active_2007}, we select instances relying on the disagreement among a committee of models.  Our committee is the set of $d=4$ previous student models plus the current -- presumably best -- student.  The disagreement is quantified by computing the proportion of committee members contradicting the current student.

\paragraph{Coreset (CS)} CS~\citep{sener_active_2018} uses an encoder to obtain the embedding representation of the new instance. Then, it calculates the cosine similarity between the embedding of the new input and the embeddings of past examples. 
If the similarity with respect to the most similar past instance $x_i$ annotated by the LLM is below a certain threshold $s$, then it requests further annotation from the LLM.
To obtain the embeddings, we average the encoder representation across tokens, as this has been proven effective in sentence embedding benchmarks~\citep{ni_sentence-t5_2021}. Similarity with previous examples has been employed in AL to encourage diversity and coverage~\citep{kim_mmr-based_2006, zeng_empirical_2019}.  GPTCache also uses the embedding representations to decide whether an incoming instance should be labelled. 

\begin{table*}
 \addtolength{\tabcolsep}{-0.3em}
\centering
\begin{tabular}{lcccc} \toprule
   & \small \textbf{ISEAR} & \small \textbf{RT-Polarity} & \small \textbf{FEVER} & \small \textbf{Openbook} \\ \midrule
\multicolumn{1}{l}{Accuracy, T5+LoRA (100 gold labels)}                     & 0.51                                & 0.85                                      & 0.53                                & 0.23                                   \\ 
\multicolumn{1}{l}{Accuracy, T5+LoRA (5000 gold labels)}                    & 0.67                                & 0.90                                      & 0.74                                & 0.68                                   \\ 
\multicolumn{1}{l}{Accuracy, LLM}                                           & 0.68                              & 0.91                                      & 0.78                                & 0.80                                   \\ \midrule
\multicolumn{1}{l}{Average margin (LLM labels)}                   & 10.0                                & 15.4                                      & 9.2                                 & 10.3                                   \\
\multicolumn{1}{l}{Average margin when wrong (LLM labels)} & 4.2                                 & 10.3                                      & 6.9                                 & 5.3                                    \\ \bottomrule
\end{tabular}
\caption{The accuracy of the LLM is similar to training the simple model with 5000 gold labels.}
\label{generated}
\end{table*}

\section{Experimental Setup}

\subsection{Datasets} 
We study the proposed setup on four classification tasks. The first two tasks have been commonly studied in AL for NLP: ISEAR~\citep{shao_universality_2015} and RT-Polarity~\citep{pang_seeing_2005}. The remaining two tasks showcase harder problems where factual knowledge acquired during pre-training of an LLM could be highly beneficial: the fact-checking dataset FEVER~\citep{thorne_fever_2018} and the question-answering dataset Openbook~\cite{mihaylov_can_2018}. We split all datasets into online and test portions (80\%-20\%, except for Openbook, as it has fewer samples). The classes are uniformly distributed for each dataset.

\paragraph{ISEAR~\citep{shao_universality_2015}} annotates
personal reports for emotion (classes: \textit{joy}, \textit{fear}, \textit{shame}, \textit{sadness}, \textit{guilt}, \textit{disgust}, \textit{anger}; 7666 examples). 

\paragraph{RT-Polarity~\citep{pang_seeing_2005}} provides sentiment polarity labels for movie reviews (classes: \textit{positive}, \textit{negative}; 10662 examples). 

\paragraph{FEVER~\citep{thorne_fever_2018}} is a fact-checking dataset (classes: \textit{true}, \textit{false}; 6612 examples) with claims that can be checked with 1-3 sentences from Wikipedia. 

\paragraph{Openbook~\citep{mihaylov_can_2018}} is a challenging question-answering dataset modelled after open book exams for assessing human understanding of a subject. Each instance consists of a multiple choice question (classes: \textit{A}, \textit{B}, \textit{C}, \textit{D}) and includes one fact that can help answer it. The full dataset consists of 5957 data points; we selected 5457 for the online set and 500 for testing. 

\subsection{Annotation by LLM}
\label{generation-datasets} 
While we are interested in the online caching scenario, to facilitate comparisons between our methods and ensure replicability in future work, we create a dataset in which we obtain LLM predictions for all data points; this dataset is then used to simulate the online setup. 

We generate soft labels using OpenAI's \texttt{text-davinci-003}, an InstructGPT-based model~\citep{ouyang_training_2022}. For each task, we design a prompt that describes the task and the possible classes. Our prompts do not contain any in-context examples (zero-shot), but we use a small part of the dataset (up to 10 examples) for prompt engineering.

On all datasets, we observe that the LLM achieves better accuracy than the smaller model trained on 5000 gold labels, suggesting that KD would be useful in these datasets (Table~\ref{generated}). In our benchmark, we store the log-probabilities of the labels. We note that the average margin for the generated labels is substantially lower when the predicted label is wrong; we observe with additional experiments that the LLM annotations are well calibrated. We release our benchmark with the generated labels to encourage further work on the neural caching problem.\footnote{\href{https://huggingface.co/datasets/guillemram97/cache_llm}{https://huggingface.co/datasets/guillemram97/cache\_llm}}   

\subsection{Experiment Details}
We run all our experiments with three random seeds, which also determine the ordering of examples; we present the average scores. For simplicity, we use a retraining frequency $f=1000$ and a constant cost per query $c(x_i)=1$. To avoid a cold-start, we train the initial student model $\mathcal{S}_{0}$ with $N=100$ (ISEAR, RT-Polarity) or $N=1000$ (FEVER, Openbook) data points from the LLM; we choose $N$ so that $\mathcal{S}_{0}$ is better than random choice. For the student model, we use $T5_{base}$~\citep{raffel_exploring_2020} as the backbone model; we freeze the model weights and add LoRA adapter layers for a parameter-efficient fine-tuning~\citep{hu_lora_2021}. 

We fine-tune the student model using the cross-entropy loss on the log probabilities assigned by the teacher in each class. We find this slightly beneficial only on FEVER in comparison to only using the most likely class (Table~\ref{soft_labels}). We split the accumulated data from the LLM into training and validation sets, and train each student from scratch for 30 epochs with early stopping with patience of five epochs. The rest of the hyperparameters can be found in Appendix~\ref{hparams}.
\section{Experiments}
\label{s.experiments}
We first present our results and then their analysis. To report accuracy across budgets, we use the corresponding Area Under the Curve (AUC) divided by the budget range, thus obtaining an average accuracy.

\subsection{Neural Caching without Student Retraining}
\label{s.neural-no-retrain}

\begin{figure*} 
    \centering
{\includegraphics[height=84pt]{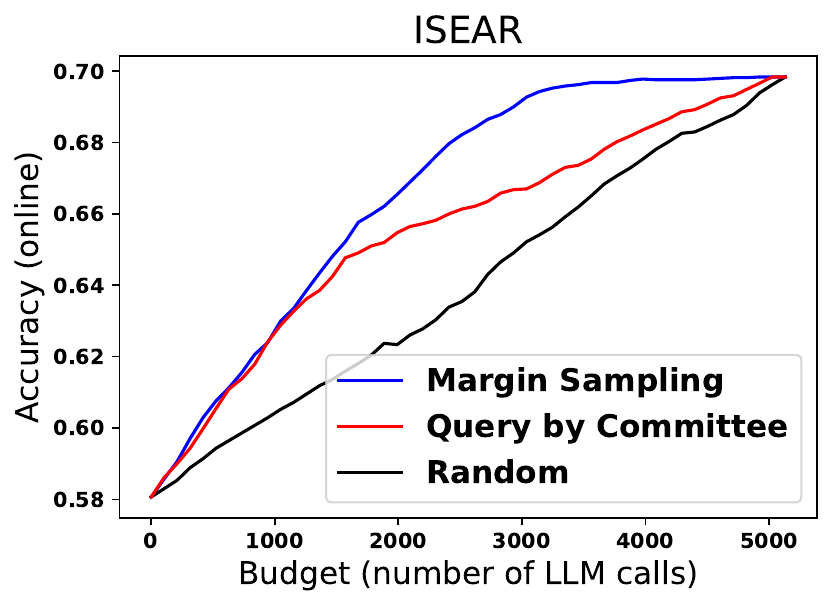}}
{\includegraphics[height=84pt]{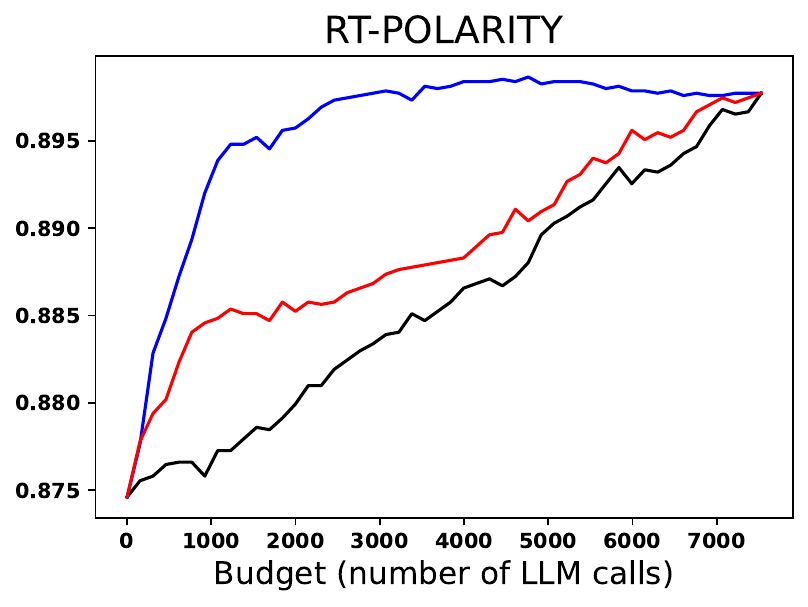}}
{\includegraphics[height=84pt]{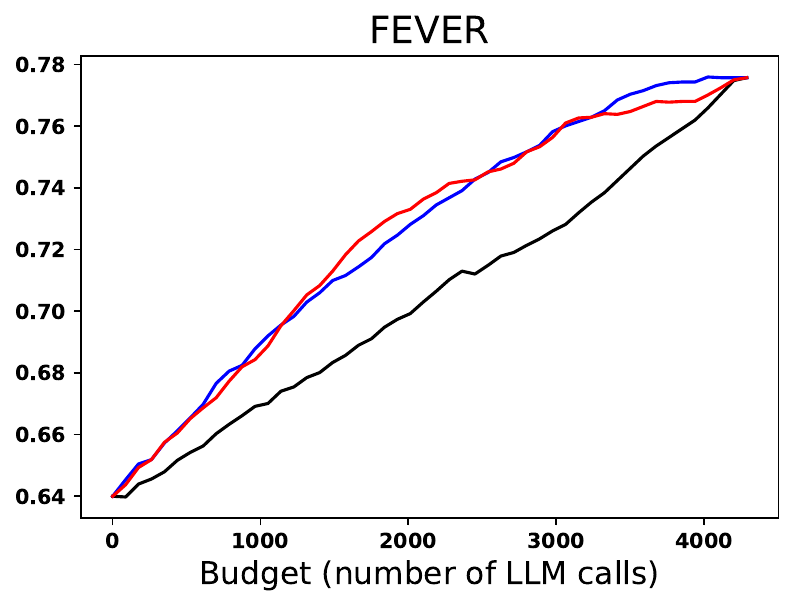}} 
{\includegraphics[height=84pt]{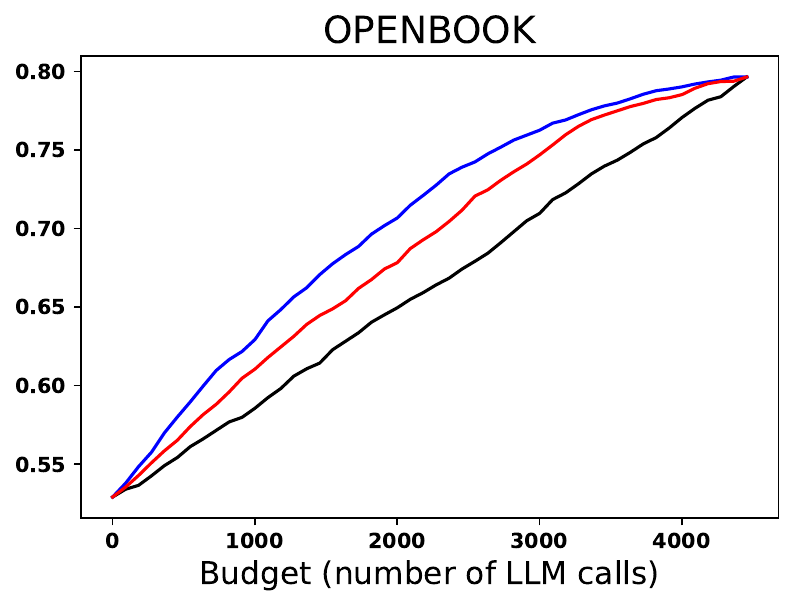}} 

    \caption{Accuracy curve with respect to budgets for neural caching without student retraining.}
    \label{online-fig-no-retrain}
\end{figure*}

\begin{table*}[t]
\centering
\begin{tabular}{lccccc} \toprule

                                        & \textbf{ISEAR} & \textbf{RT-Polarity} & \textbf{FEVER} & \textbf{Openbook} & \textbf{Average} \\ 

\midrule

\multicolumn{1}{l}{Random}             & 0.640          & 0.886                & 0.704          & 0.662             & 0.723            \\ 
\multicolumn{1}{l}{Margin Sampling}    & \textbf{0.666}          & \textbf{0.896}                & \textbf{0.725}          & \textbf{0.703 }            & \textbf{0.748}            \\ 
\multicolumn{1}{l}{Query by Committee} & 0.656          & 0.889                & \textbf{0.725}          & 0.687             & 0.739            \\\bottomrule

\end{tabular}
\caption{Online accuracy (AUC) for neural caching with no student retraining.}
\label{online-backoff-table-bona}
\end{table*}

We first study a simplified version of neural caching, where the student model is not retrained on new data points. This is a practical scenario, as retraining creates extra overhead for the application provider (e.g., consider a setting where the student is run on a portable device, which is not powerful enough to support retraining). 

We adapt the AL instance selection criteria in the following way. Given a criterion $C$, we calculate the respective values from the previous outputs of the student and call this list the history $\hat{C}$. If we have a remaining budget $b$ and $n$ remaining online instances, we use as a threshold for an incoming instance the $\frac{b}{n}$-th percentile of the history $\hat{C}$. The best possible scenario would imply having oracle threshold values for each budget (i.e. as if we had access to the full dataset offline). However, in additional experiments, we found that the above rule yields very similar scores. 

To use QBC in this setup, we simulate that we have four previous students trained on subsets of the data. For example, if the student is trained on $N=1000$ examples, the previous students are trained on 900, 800, 700, and 600 data points, respectively.
We find that MS yields results very similar to PE and that Coreset is similar to Random. To ease visualising the results, here we omit PE and Coreset.

Table~\ref{online-backoff-table-bona} and Figure~\ref{online-fig-no-retrain} contain the results when we train the initial student with $N=1000$ datapoints annotated by the LLM. Our experiments with different initial budgets $N$ yield similar results (Table~\ref{online-backoff-table-extensive} in Appendix). 

We find that MS is the best-performing method on all datasets and across all the initial student models, followed by QBC, which outperforms the baseline of random selection. Given the simplicity of both methods, these results make a strong case for using AL-based selection methods, especially MS. Unlike QBC, MS does not require storing multiple models and performing inference with each of them. 

\begin{figure} 
    \centering
\includegraphics[width=\linewidth]{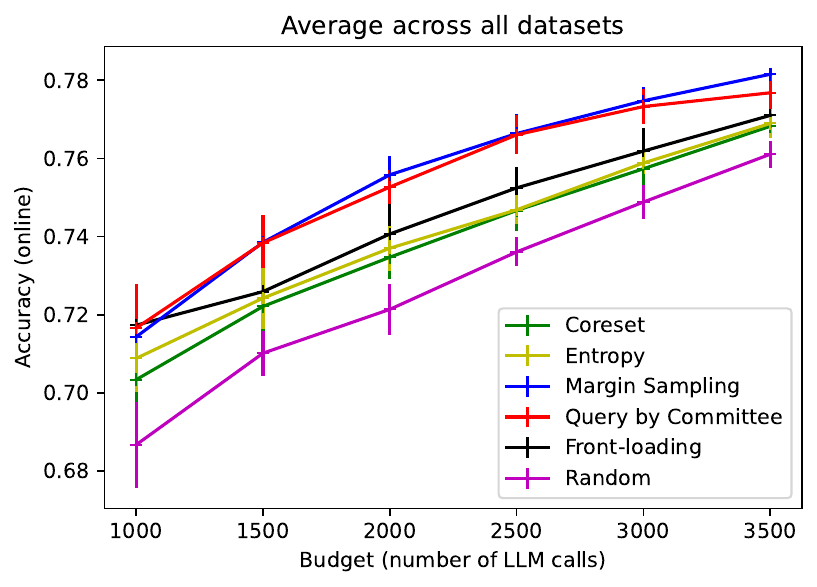}
    \caption{Accuracy curve with respect to budgets, in the neural caching problem with student retraining. Error lines indicate variance. We have averaged results across the four datasets.}
    \label{online-retrain-avg}
\end{figure}

\begin{figure*}
    \centering
    {\includegraphics[width=200pt, height=150pt]{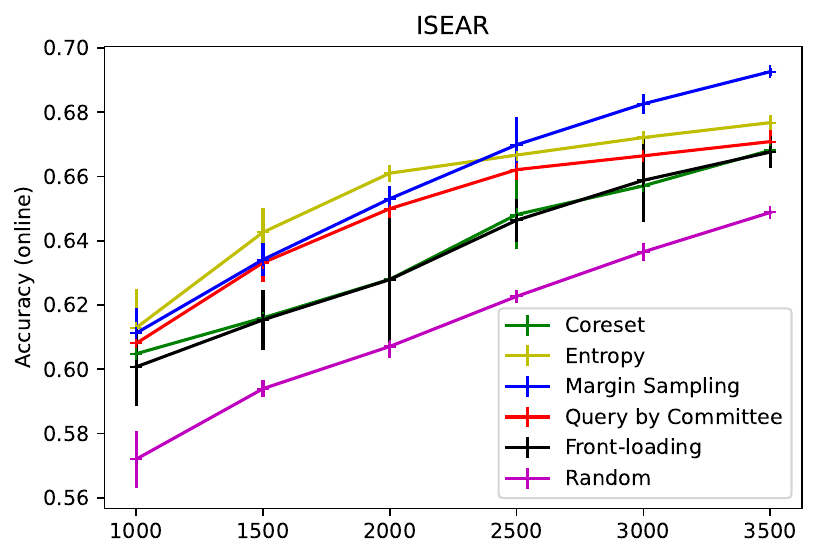}}
    {\includegraphics[width=200pt, height=150pt]{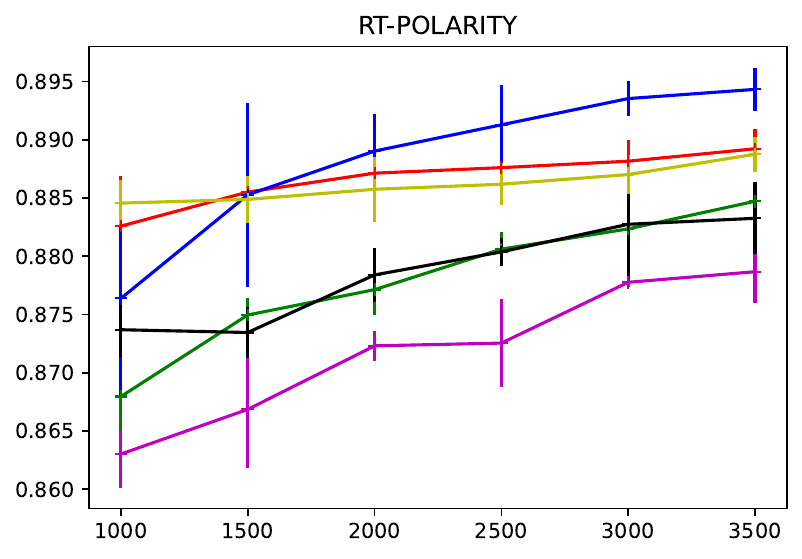}}
    {\includegraphics[width=200pt, height=160pt]{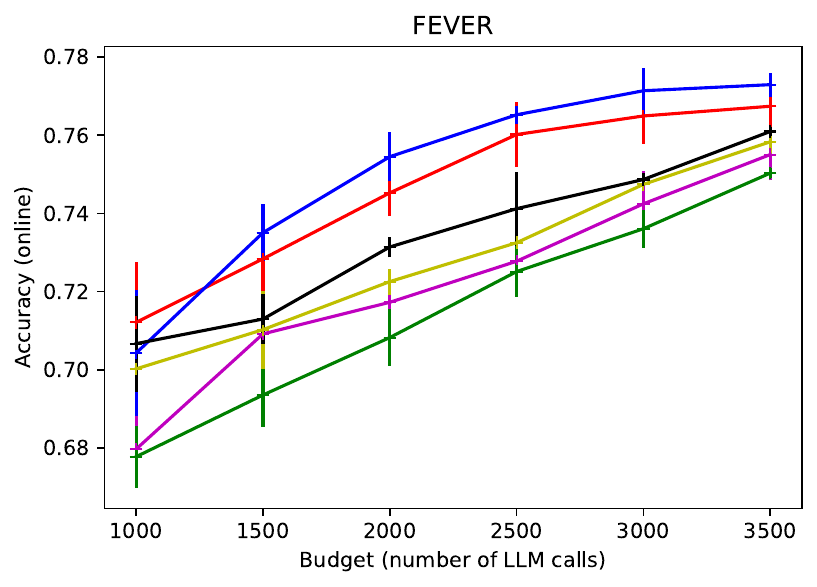}}
    {\includegraphics[width=200pt, height=160pt]{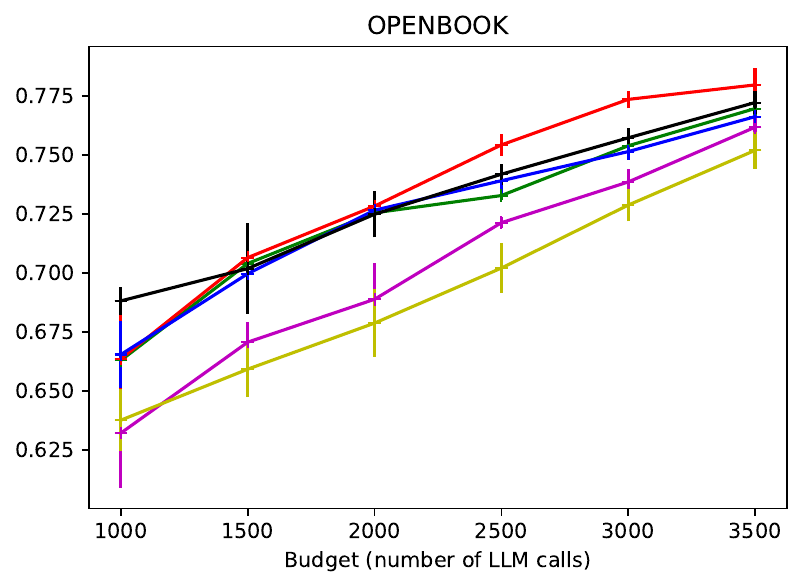}} 
    \caption{Accuracy curve with respect to budgets, in the neural caching problem with student retraining. Error lines indicate variance.}
    \label{online-fig}
\end{figure*}

\begin{table*}
\centering
\begin{tabular}{lccccc} \toprule
\textbf{}                                & \textbf{ISEAR} & \textbf{RT-Polarity} & \textbf{FEVER} & \textbf{Openbook} & \textbf{Average} \\ \midrule
\multicolumn{1}{l}{Random}             & 0.614          & 0.872                & 0.723          & 0.703             & 0.728            \\ 
\multicolumn{1}{l}{Front-loading}      & 0.637          & 0.879                & 0.734          & 0.731             & 0.745            \\ 
\multicolumn{1}{l}{Coreset}            & 0.637          & 0.878                & 0.715          & 0.726             & 0.739            \\ 
\multicolumn{1}{l}{Entropy}            & 0.657          & 0.886                & 0.728          & 0.693             & 0.741            \\ 
\multicolumn{1}{l}{Margin Sampling}    & \textbf{0.658}          & \textbf{0.889}                & \textbf{0.753}          & 0.726             & \textbf{0.757}           \\ 
\multicolumn{1}{l}{Query by Committee} & 0.650          & 0.887                & 0.748          & \textbf{0.737}             & 0.755            \\ \bottomrule
\end{tabular}
\caption{Online accuracy (AUC) for neural caching with student retraining.}
\label{retraining}
\end{table*}

\begin{table*} 
\centering
\begin{tabular}{cccccc} \toprule
& \textbf{ISEAR} & \textbf{RT-Polarity} & \textbf{FEVER} & \textbf{Openbook} & \textbf{Average} \\ \midrule
\multicolumn{1}{l}{Front-loading}      & 0.598          & 0.879                & 0.686          & \textbf{0.647}             & 0.702            \\
\multicolumn{1}{l}{Coreset}            & 0.599          & 0.879                & 0.680          & 0.641             & 0.700            \\ 
\multicolumn{1}{l}{Entropy}            & 0.608          & \textbf{0.885}              & 0.682          & 0.647             & 0.705            \\ 
\multicolumn{1}{l}{Margin Sampling}    & \textbf{0.609}          & 0.884                & 0.678          & 0.634             & 0.701            \\ 
\multicolumn{1}{l}{Query by Committee} & \textbf{0.609}          & 0.882                & \textbf{0.687}          & 0.646             & \textbf{0.706}           \\ \bottomrule
\end{tabular}
\caption{Final accuracy (AUC) of the last student model for neural caching with student retraining.}
\label{retraining-test}
\end{table*}

\subsection{Neural Caching with Student Retraining}
\label{s.online-backoff-retrain}
We now turn to the complete setup proposed in Section~\ref{caching}, in which the selected instances are used to retrain a student model with some periodicity. This creates the incentive to spend the budget early to get a more proficient student model as soon as possible. To observe this effect, we include a random baseline with a uniform sampling rate. This suggests waiting longer for informative examples to arrive counterweights the benefits of getting a strong student as quickly as possible. We select thresholds to encourage spending more of the budget early on (see Appendix \ref{thresholds}). 

We show the results averaged across all datasets in Figure~\ref{online-retrain-avg} and per-dataset in Figure~\ref{online-fig}. We observe that both MS and QBC considerably outperform the other methods. The embedding-based strategy (Coreset) does badly in all the studied setups. Table~\ref{retraining} summarises the results. 

\subsection{Analysis}

\begin{figure}
    \centering
\includegraphics[height=75pt]{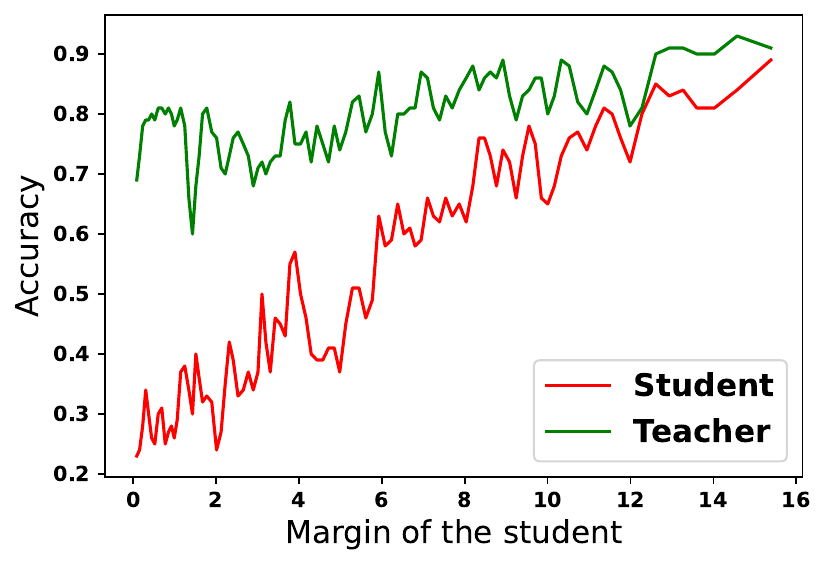}{}\includegraphics[height=75pt]{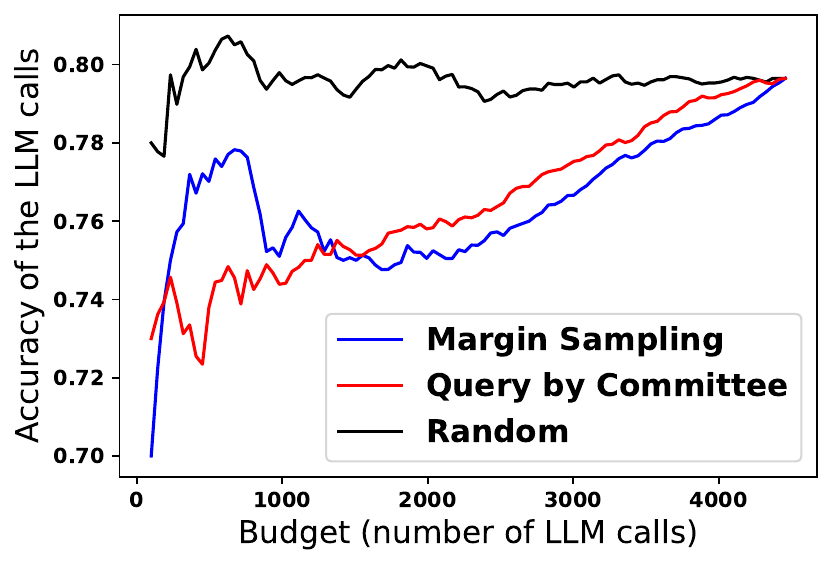}
    \caption{On the left, we order data points by their margin and plot the accuracy of their respective labels generated by the student and teacher. We observe that the greatest advantage of using the labels from the teacher comes with low margins. On the right, the accuracy of the labels generated by the LLM calls in neural caching with no student retraining. We observe that MS and QBC are more likely to generate wrong labels. We focus on Openbook for both plots.}
    \label{backoff-off-bad}
\end{figure}

\paragraph{Hard examples with noisy labels.} We have observed in our experiments that prioritising harder instances for teacher annotation leads to clear gains in online accuracy. However, as discussed in the introduction, LLM accuracy may be significantly affected by the increased `complexity' of an example, which can inflate the proportion of noisy annotations in the data on which the student is trained (see Figure~\ref{backoff-off-bad}). This problem is known in KD as \emph{confirmation bias}~\citep{arazo_pseudo-labeling_2020,liu_certainty_2021}.  Previous results from offline KD suggest that this type of confirmation bias can be mitigated by avoiding the hardest instances~\cite{baykal_robust_2023}, improving the chances that the teacher model makes a correct prediction. However, we observe that the most significant advantage of the LLM with respect to the student in terms of accuracy lies in these samples that are deemed hard by the student (leftmost part of the plot in Figure~\ref{backoff-off-bad}); since we are optimising the online accuracy, the trade-off between providing hard or correct labels may be different in our online case than in the offline scenario.
Given the above, we hypothesise that MS and QBC would be more negatively affected by the confirmation bias than front-loading, which does not prioritise hard examples. To test this hypothesis, we designed an experiment to put an upper bound on the effect of wrong LLM annotations. For each strategy, we only retrain the student model on correct labels, simulating an oracle that discards incorrect examples. Table~\ref{oracle-1} shows the absolute improvements in the online and final accuracy with respect to the values obtained without the oracle (Table~\ref{retraining} and~\ref{retraining-test}). We observe moderate absolute improvements, but surprisingly MS and QBC do not seem to improve more than front-loading, suggesting that the hypothesis is wrong and that the impact of confirmation bias is somewhat limited and - what is surprising - similar across strategies. 

As an additional test, we analyse the subset of test examples where the teacher is incorrect. If confirmation bias is a major issue for MS and QBC than for front-loading, we would expect that they are more prone to reproducing the teacher's errors. Again, we do not find any substantial differences between these two strategies vs front-loading (Table \ref{conf-bias}).

\paragraph{Online accuracy vs.\ final accuracy.} Taking a look at the accuracy of the final student (Table~\ref{retraining-test}), we observe that it is generally consistent with the online accuracy (Table~\ref{retraining}). However, MS has a low \textit{final} accuracy on FEVER while having the best \textit{online} accuracy on that dataset, confirming that in some tasks, calling the LLM to obtain labels for hard examples may improve the online accuracy while not necessarily improving the student.  This result emphasises the differences between our set-up and the setting normally studied in AL.

 \addtolength{\tabcolsep}{-0.3em}
\begin{table} 
\centering
\begin{tabular}{lcc}
\toprule
                   &  \footnotesize $\Delta$Online
                    &  \footnotesize $\Delta$Final 
                   \\ \midrule

Front-loading      &  \footnotesize 0.009             & \footnotesize 0.019            \\
Margin Sampling    & \footnotesize 0.008             & \footnotesize 0.022            \\
Query by Committee & \footnotesize 0.008             & \footnotesize 0.018 \\ \bottomrule
\end{tabular} 
\caption{Absolute improvements for the online and final accuracy using an oracle that allows us to discard instances with wrong labels from the LLM, averaged across datasets. The improvements are with respect to values from Table~\ref{retraining} and~\ref{retraining-test}.}
\label{oracle-1}
\end{table}

\subsection{Robustness of the Findings}

\paragraph{Vary initial training ($\mathcal{S}_{0}$).}
We study the effect of the quantity of LLM-annotated data on which the first student model is trained, focusing on the setup with retraining (Table~\ref{ablation-budget-init}). We consider the two more challenging tasks, FEVER and Openbook.  
We find that QBC performs best overall, and the performance of MS is more sensitive to the initial budget.
This observation suggests that determining the decision criteria for transitioning from a front-loading regime to MS poses a relevant question, which we leave for future exploration.

\begin{table} 
\resizebox{\columnwidth}{!}{%
\centering
\begin{tabular}{lcccccc}
\toprule
\textbf{}          & \multicolumn{3}{c}{\textbf{Openbook}}            & \multicolumn{3}{c}{\textbf{FEVER}}               \\
\cmidrule(lr){2-4}\cmidrule(lr){5-7}
                   & $N$=1000         & $N$=2000         & $N$=3000         & $N$=500          & $N$=1000         & $N$=1500         \\ \midrule
Front-loading      & 0.731          & 0.769          & 0.751          & 0.716          & 0.734          & 0.734          \\
Margin Sampling    & 0.726          & 0.777          & 0.764          & 0.718          & \textbf{0.753} & 0.751          \\
Query by Committee & \textbf{0.737} & \textbf{0.786} & \textbf{0.779} & \textbf{0.722} & 0.748          & \textbf{0.755} \\
\bottomrule
\end{tabular}
}
\caption{Online accuracy (AUC) of different selection
criteria with different initial student models $\mathcal{S}_0$.}
\label{ablation-budget-init}
\end{table}

\paragraph{Higher retraining frequency $f$.}
We repeat neural caching experiments, setting this time a higher frequency of retraining $f=100$; this results in much longer runs as the student model has been retrained an order of magnitude more times. Table~\ref{freq} shows the results. We observe that results are consistent and very similar to those of a lower frequency of retraining (Table~\ref{retraining}).

\begin{table} 
\resizebox{\columnwidth}{!}{%
\centering
\begin{tabular}{lccccc}
\toprule
\textbf{}          & \textbf{ISEAR} & \textbf{RT-Polarity} & \textbf{FEVER} & \textbf{Openbook} & \textbf{Average} \\ \midrule
Front-loading             & 0.637          & 0.879                & 0.734          & 0.731             & 0.745            \\
Margin Sampling    & \textbf{0.661}          & \textbf{0.892}                & 0.750          & 0.728             & 0.758            \\
Query by Committee & 0.657          & 0.890                & \textbf{0.751}          & \textbf{0.740}             & \textbf{0.759}          \\ \bottomrule
\end{tabular}
}
\caption{Online accuracy (AUC) for neural caching with retraining frequency $f=100$.}
\label{freq}
\end{table}

\section{Conclusions}
In this work, we have studied how instance selection criteria from AL behave when they are used to decide in real time whether we should perform an LLM call or use a student model that has been trained on previous LLM predictions. In the scenario where we are not retraining the student model, Margin Sampling performs the best, across different datasets. In the scenario where we retrain the student model with some time periodicity, Query by Committee is the most robust option. In our experiments we observe that, while Margin Sampling outperforms the front-loading baseline on harder tasks, it is more sensitive to the initial budget spent to train the student model $\mathcal{S}_{0}$. 

We did not find the embedding-based strategy effective; it is the only LLM caching approach which is known to be adopted by practitioners (e.g., GPTCache). We believe these types of strategies could be useful in certain contexts, e.g. multiple near-identical calls to an LLM, the scenario which has not been the focus of this work.\footnote{FEVER does contain paraphrases or statements entailing each other but these constitute only a small fraction of the dataset.} 

Our results suggest that (i) there is room for smart LLM query allocation in the context of continuously distilling an LLM into a student model and (ii) previous literature in active learning can transfer well to this setup. This is, to our knowledge, the first work that leverages online knowledge distillation, which we believe could play a key role in caching LLMs and saving unnecessary calls to expensive models. 

In this work, we focused on a stationary (i.i.d.) stream of requests. In practice, the distribution of requests is likely to change over time~\cite{cacciarelli_survey_2023}. As suggested by the online AL literature~\cite{DBLP:conf/sdm/BifetG07}, this should further increase the gap between the AL-based approaches and static strategies, e.g., front-loading. In those cases, we would expect improvements in both online and final accuracy. We leave this investigation for future work.

\bibliography{references}
\bibliographystyle{acl_natbib}

\onecolumn

\appendix
\appendix
\section{Experimental details and hyperparameters}
\label{hparams}
\paragraph{Student model}
We use the T5 implementation from Huggingface's \texttt{transformers} library. We use LoRA adapters~\citep{hu_lora_2021}, as they have been considered one of the most
parameter-efficient architectures in few-shot settings~\citep{liu_few-shot_2022}. Following~\citet{ponti_combining_2022}, we add a LoRA adapter to the query, key, value and output weights in each self-attention layer of T5. We set the LoRA rank to $r=16$, and the scaling to $\alpha=0.25$. \\
We use learning rate $\eta=5\cdot10^{-4}$, training batch size $m=16$ and weight decay $\lambda=0.01$. We validate this hyperparameter choice based on experiments using the soft labels from the teacher. 
\paragraph{Adaptation of strategies}
For Entropy, we normalise before computing it by applying a softmax over the classes. 
\paragraph{Reporting of results} 
In order to report accuracy across budgets, we use the corresponding Area Under the Curve (AUC) divided by the budget range. By budget range, we refer to the biggest budget minus the smallest one for that task.

\subsection{Threshold values}
\label{thresholds}
To encourage an early expense of the budgets in the setting with student retraining, we have selected threshold values to ensure initially a higher proportion of calls for LLM annotation (PE=0.5, MS=5, QBC=4, CS=0.9); we have selected these values so that the first student model selects at least $50\%$ of instances for LLM annotation on RT-Polarity. However, we observe very similar results when we use the empirical threshold from Section \ref{s.neural-no-retrain}.

\subsection{Labels from the LLM}
We use a budget for LLM annotation of \$200. All the labels are obtained during May 2023. Since the OpenAI API can only return up to the five most likely tokens, we add a bias $b=100$ to the tokens that represent each class:
\begin{itemize}
    \item ISEAR: '\textit{ joy}', '\textit{ fear}', '\textit{ anger}', '\textit{ sadness}', '\textit{ disgust}', '\textit{ shame}', '\textit{ guilt}'
    \item  RT-POLARITY: '\textit{ positive}', '\textit{ negative}'
    \item FEVER: '\textit{ true}', '\textit{ false}'
    \item OPENBOOK: '\textit{ A}', '\textit{ B}', '\textit{ C}', '\textit{ D}'
\end{itemize}
If a class is not among the five most likely tokens, it gets assigned in our experiments a log probability of -100.

\section{Additional results}

\subsection{Neural caching with no student retraining}

\begin{table*}[t]
\centering
\begin{tabular}{cccccc}\toprule
    & \textbf{ISEAR} & \textbf{RT-Polarity} & \textbf{FEVER} & \textbf{Openbook} & \textbf{Average} \\ \midrule
Random ($N$=500)            & 0.629          & 0.882                & 0.679          & 0.567             & 0.689            \\ 
Margin Sampling ($N$=500)    & \textbf{0.656}          & \textbf{0.895}                & \textbf{0.698}          & \textbf{0.587}             & \textbf{0.709}            \\ 
Query by Committee ($N$=500) & 0.644          & 0.887                & 0.693          & 0.568             & 0.698            \\ \midrule
Random ($N$=1000)            & 0.640          & 0.886                & 0.704          & 0.662             & 0.723            \\ 
Margin Sampling ($N$=1000)    & \textbf{0.666}          & \textbf{0.896}                & \textbf{0.725}          & \textbf{0.703}             & \textbf{0.748}            \\ 
Query by Committee ($N$=1000) & 0.656          & 0.889                & \textbf{0.725}          & 0.687             & 0.739            \\ \midrule
Random ($N$=2000)             & 0.652          & 0.884                & 0.724          & 0.729             & 0.747            \\ 
Margin Sampling ($N$=2000)    & \textbf{0.673}          & \textbf{0.896}               & \textbf{0.751}          & \textbf{0.764}             & \textbf{0.771}            \\
Query by Committee ($N$=2000) & 0.667          & 0.891                & 0.745          & 0.760             & 0.766            \\ \midrule
Random ($N$=3000)             & 0.648          & 0.885                & 0.738          & 0.734             & 0.752            \\ 
Margin Sampling ($N$=3000)    & \textbf{0.669}          & \textbf{0.895}                & 0.757          & 0.767             & \textbf{0.772}            \\ 
Query by Committee ($N$=3000) & 0.665          & 0.890                & \textbf{0.758}          & \textbf{0.773}             & 0.771            \\ \bottomrule
\end{tabular}
\caption{Online accuracy (AUC) for neural caching with no retraining.}
\label{online-backoff-table-extensive}
\end{table*}

We observe that Margin Sampling is the best-performing method on all datasets and across all the initial student models, followed by Query by Committee and outperforming the baseline of random selection (Table~\ref{online-backoff-table-extensive}). The gap between Margin Sampling and the baseline widens as we have a better initial student. 

\subsection{Soft labels}
We conduct experiments to study the effect of using soft labels (using the logprobabilities for each class from the LLM) or hard labels (only using the first class from the LLM). To do this, we train a student model on multiple budgets and obtain the final accuracy. We observe this has some gains in FEVER (Table~\ref{soft_labels}). 
\begin{table*}  
\centering
\begin{tabular}{ccccccc}

\toprule
    & \textbf{ISEAR} & \textbf{RT-Polarity} & \textbf{Openbook} & \textbf{FEVER} & \textbf{Average} \\ \midrule
Soft labels & \textbf{0.598}          & \textbf{0.880}                & \textbf{0.617}             & \textbf{0.670}          & \textbf{0.691}            \\ 
Hard labels & \textbf{0.598}          & 0.879                & 0.616             & 0.659          & 0.688            \\ \bottomrule
\end{tabular}
\caption{Final accuracy (AUC) of the last student model, taking either soft or hard labels from the LLM.}
\label{soft_labels}
\end{table*}

\subsection{Effect of confirmation bias in neural caching with retraining}
To study the confirmation bias, we select the samples from the test dataset where the LLM produces a wrong answer. If the model performance is affected by the noise of the labels it was trained on, it is expected it will reproduce the mistakes of the LLM; therefore, we would expect that it will have a lower score in this subset of the test dataset. We do not find that Margin Sampling and Query by Committee have lower performance than front-loading in this subset of the dataset (Table~\ref{conf-bias}).  \\

\begin{table}  
\centering
\begin{tabular}{cccc} \toprule
                   & \textbf{ISEAR} & \textbf{FEVER} & \textbf{Openbook} \\ \midrule
Front-loading      & 0.171          & 0.513          & 0.339             \\
Margin Sampling    & 0.180          & 0.497          & 0.340             \\
Query by Committee & 0.178          & 0.512          & 0.344          \\ \bottomrule  
\end{tabular}
\caption{Accuracy (AUC) over the subset of the test dataset where the LLM produces wrong labels for the last student model for neural caching with student retraining.
}
\label{conf-bias}
\end{table}

\section{Prompts used}
The following are the prompts we used when calling the LLM. We have marked in blue one example, and in red the expected answer.
\begin{itemize}
    \item \textbf{ISEAR}: This is an emotion classification task. Only answer one of: 'joy', 'fear', 'anger', 'sadness', 'disgust', 'shame', 'guilt'. \\ INPUT: \color{blue} During the period of falling in love, each time that we met and especially when we had not met for a long time. \color{black} \\
OUTPUT: \color{red} joy
\color{black}
    \item 
\textbf{RT-Polarity}: This is a sentiment classification task for movie reviews. Only answer either 'positive' or 'negative'.\\
INPUT: \color{blue} if you sometimes like to go to the movies to have fun , wasabi is a good place to start .  \color{black}\\
OUTPUT: \color{red} positive \color{black}
    \item \textbf{FEVER}: This is a fact-checking task. Only answer either 'true' or 'false'.\\
    INPUT: \color{blue} On June 2017, the following claim was made: Jeb Bush is former President George H. W. Bush's daughter. Q: Was this claim true or false? \\
    \color{black}
    OUTPUT: \color{red} false 
    \color{black}
    \item \textbf{Openbook}: This is a multiple-choice test. You are presented a fact and a question. Only answer one letter, producing no more output.\\
\color{blue}
FACT: the sun is the source of energy for physical cycles on Earth \\
QUESTION: The sun is responsible for \\
A: puppies learning new tricks \\
B: children growing up and getting old \\
C: flowers wilting in a vase \\
D: plants sprouting, blooming and wilting \color{black} \\
OUTPUT: \color{red} D \color{black}

\end{itemize}

\end{document}